\documentclass[a4paper,11pt]{amsart}%

\usepackage{cite}

   \usepackage{graphicx}
    \usepackage{tikz}

\usepackage{amsmath}
\usepackage{amsfonts}
\usepackage{amssymb}
 \usepackage{multirow}

\usepackage{fullpage}

\usepackage{array}

  \usepackage[caption=false,font=normalsize,labelfont=sf,textfont=sf]{subfig}

\usepackage{stfloats}

\begin{document}

\title{Variational Bayesian Inference For A Scale Mixture Of Normal Distributions Handling Missing Data}

\author{Guillaume~Revillon,
        Ali~Mohammad-Djafari
        and~Cyrille~Enderli.} 
\thanks{M. Revillon is with Laboratoire des Signaux Syst\`{e}mes, Centrale Sup\'{e}lec, Universit\'{e} Paris Saclay and Thales Syst\`{e}mes A\'{e}roport\'{e}s, France. Email : guillaume.revillon@l2s.centralesupelec.fr}
\thanks{M. Mohammad-Djafari is with Laboratoire des Signaux Syst\`{e}mes, Centrale Sup\'{e}lec, Universit\'{e} Paris Saclay and CNRS, France. Email : Ali.mohammad-Djafari@l2s.centralesupelec.fr}
\thanks{M. Enderli is with Thales Syst\`{e}mes A\'{e}roport\'{e}s, France. Email : cyrille-jean.enderli@fr.thalesgroup.com}


\maketitle

\begin{abstract}
In this paper, a scale mixture of Normal distributions model is developed for classification and clustering of data having outliers and missing values. The classification method, based on a mixture model, focuses on the introduction of latent variables that gives us the possibility to handle sensitivity of model  to outliers and  to  allow  a less restrictive  modelling of missing data. Inference is processed through a Variational Bayesian Approximation and a Bayesian treatment is adopted for model learning, supervised classification and clustering. 
\end{abstract}


\section{Introduction}
The main objective in this paper is to define a classification/clustering framework that handles both outliers and missing values. 
Gaussian mixture models \cite{Quandt1978} (GMM) are the most famous mixture models for continuous data and have been widely used for decades. Indeed, as weighted sums of Gaussian distributions GMMs  benefit from attractive Gaussian properties, then dependency between features can easily be modelled through a multivariate Gaussian distribution in order to infer on missing values. However, a major limitation of GMMs is their lack of robustness to outliers that can lead to over-estimate the number of clusters since they use additional components to capture the tails of the distributions \cite{Svensen2005}. To fill that lack, we proposed to use a scale mixture of Normal distributions \cite{Andrews1974} in a Bayesian framework. The main advantages of this model is that the model accounts for the uncertainties of variances and covariances since the associated marginal distributions are heavy-tailed \cite{Archambeau2007}. Exact inference in that Bayesian approach is unfortunately intractable and a Variational Bayesian (VB) inference \cite{Waterhouse1996} is used to estimate the posterior distribution. The proposed model is explained in Section \ref{section:2} and inference procedure is derived in Section \ref{section:3}.

\section{Model}
In this section,  the standard  GMM is briefly presented  as a hierarchical latent variable model  before introducing  missing values and outliers modelling. At last, the  proposed  model is developed. 

\label{section:2}

\subsection{Latent variable model}
A GMM  \cite{Jordan1994} is a natural  framework for classification and  clustering. It can be formalized  as : 
        
        		\begin{equation}\label{mixture}
        			p(\boldsymbol{x}|\boldsymbol{\Theta},\mathcal{K})=\sum\limits_{k\in\mathcal{K}}a_k\mathcal{N}(\boldsymbol{x}|\boldsymbol{\mu}_k,\boldsymbol{\Sigma}_k)~,
        		\end{equation}

		\begin{flushleft}
			where $\boldsymbol{x}\in\mathbb{R}^d$ is an observation, $\mathcal{K}=\{1,\ldots,K\}$ is a finite and known set of clusters and $\boldsymbol{\Theta}=(\boldsymbol{a},(\boldsymbol{\mu}_k,\boldsymbol{\Sigma}_k)_{k\in\mathcal{K}})$, with $\boldsymbol{a}=[a_1,\ldots,a_K]'$, stands for parameters. Moreover, $ \boldsymbol{\mu}_k$ and $\boldsymbol{\Sigma}_k$ are respectively the mean and the covariance matrix of the $k^{th}$  component distribution  with a weight $a_k$ where $a_k\geq 0$ and $\sum_{k\in\mathcal{K}}a_k=1$.
		\end{flushleft}
				
		The GMM can be formalized as a latent model since the component label associated to each data point is unobserved. To this end, a categorical variable $z\in \mathcal{K}$ can be considered to describe the index of the component distribution generating the observation variable $\boldsymbol{x}$.  Then, the mixture distribution  (\ref{mixture}) is expressed  as 
	\begin{equation}\label{latent}
      		p(\boldsymbol{x}|\boldsymbol{\Theta},\mathcal{K})=\sum\limits_{z\in \mathcal{K}}p(\boldsymbol{x}|z,\boldsymbol{\Theta},\mathcal{K})p(z|\boldsymbol{\Theta},\mathcal{K})\;,
      	\end{equation}
	\begin{flushleft}
         	where
	\end{flushleft}
	\begin{align}\label{latent2}
		 	&p(\boldsymbol{x}|z,\boldsymbol{\Theta},\mathcal{K})=\prod_{k\in\mathcal{K}}\mathcal{N}(\boldsymbol{x}|\boldsymbol{\mu}_k,\boldsymbol{\Sigma}_k)^{\delta_{z}^k}\;,\\
			&p(z|\boldsymbol{\Theta},\mathcal{K})=\mathcal{C}\textit{at}(z|\boldsymbol{a})=\prod_{k\in\mathcal{K}}a_k^{\delta_{z}^k}\;
	\end{align}
	\begin{flushleft}
	and $\delta_{z}^k$ denotes the Kronecker symbol which is $1$ if $z=k$ and  $0$ otherwise.
	\end{flushleft}

	\subsection{Missing values}
	Missing values can be handled by decomposing the features vector $\boldsymbol{x} \in \mathbb{R}^d$ into observed features $\boldsymbol{x}^{\text{obs}}\in\mathbb{R}^{d_{\text{obs}}}$ and missing features modelled by a latent variable $\boldsymbol{x}^{miss}\in\mathbb{R}^{d_{\text{miss}}}$ such that $1\le d_{\text{obs}} \le d $ and $d_{\text{miss}}=d-d_{\text{obs}}$. Reminding that conditionally to its index cluster the features vector $\boldsymbol{x}$ is Gaussian distributed		
		
		\begin{equation*}
		\boldsymbol{x}=\begin{pmatrix} 
			\boldsymbol{x}^{\text{miss}}\\
			\boldsymbol{x}^{\text{obs}} 
		\end{pmatrix} | z=k
		\sim\mathcal{N}\left(\boldsymbol{\mu}_k = \begin{pmatrix} 
			\boldsymbol{\mu}_k^{\text{miss}}\\
			\boldsymbol{\mu}_k^{\text{obs}} 
		\end{pmatrix},
		\boldsymbol{\Sigma}_k=\begin{pmatrix} 
			\boldsymbol{\Sigma}_k^{\text{miss}} & \boldsymbol{\Sigma}_k^{\text{cov}} \\
			\boldsymbol{\Sigma}_k^{\text{cov}'}& \boldsymbol{\Sigma}_k^{\text{obs}}
		\end{pmatrix}\right),
		\end{equation*}
		
		\begin{flushleft}
		the latent variable  $\boldsymbol{x}^{\text{miss}}$ can be expressed as a Gaussian distributed variable conditionally to $z$ such that 
		\end{flushleft}
		
		\begin{equation}\label{missing}
			p(\boldsymbol{x}^{miss}|\boldsymbol{x}^{obs},z,\boldsymbol{\Theta},\mathcal{K}))=\prod_{k\in\mathcal{K}} \mathcal{N}\left(\boldsymbol{x}^{miss}|\boldsymbol{\epsilon}_k^{\text{miss}},\boldsymbol{\Delta}_k^{\text{miss}}\right)^{\delta_z^k}
		\end{equation}
		
		\begin{flushleft}
			where
		\end{flushleft}
		
		\begin{align*}
			&\boldsymbol{\epsilon}_k^{\text{miss}}=\boldsymbol{\mu}_k^{\text{miss}}+\boldsymbol{\Sigma}_k^{\text{cov}}\boldsymbol{\Sigma}_k^{\text{obs}^{-1}}(\boldsymbol{x}^{\text{obs}}-\boldsymbol{\mu}_k^{\text{obs}} )\\
			&\boldsymbol{\Delta}_k^{\text{miss}}= \boldsymbol{\Sigma}_k^{\text{miss}}-\boldsymbol{\Sigma}_k^{\text{cov}}\boldsymbol{\Sigma}_k^{\text{obs}^{-1}} \boldsymbol{\Sigma}_k^{\text{cov}'}\;.
		\end{align*}
		
	Then, the joint distribution of $(\boldsymbol{x}^\text{miss},\boldsymbol{x}^\text{obs})$ is derived from (\ref{missing}) such that 
	
			\begin{equation*}
				p(\boldsymbol{x}^\text{miss},\boldsymbol{x}^\text{obs}|z,\mathcal{K})=\prod_{k\in\mathcal{K}} \left[ \mathcal{N}\left(\boldsymbol{x}^{miss}|\boldsymbol{\epsilon}_k^{\text{miss}},\boldsymbol{\Delta}_k^{\text{miss}}\right)p_k\left(\boldsymbol{x}^{\text{obs}}\right)\right]^{\delta_z^k}
			\end{equation*}
			
		\begin{flushleft}
		with
		\end{flushleft}
		
		\begin{align*}
			&p_k\left(\boldsymbol{x}^{\text{obs}}\right)=\mathcal{N}\left(\boldsymbol{x}^{\text{obs}}|\boldsymbol{\mu}_k^{\text{obs}},\boldsymbol{\Delta}_k^{\text{obs}}\right)\;,\\
			&\boldsymbol{\Delta}_k^{\text{obs}}= \left(\boldsymbol{\Sigma}_k^{\text{obs}^{-1}}+2\times\boldsymbol{\Sigma}_k^{\text{obs}^{-1}}\boldsymbol{\Sigma}_k^{\text{cov}'}\left(\boldsymbol{\Delta}_k^{\text{miss}}\right)^{-1}\boldsymbol{\Sigma}_k^{\text{cov}}\boldsymbol{\Sigma}_k^{\text{obs}^{-1}}\right)^{-1}\hspace{-8pt}.
		\end{align*}

		\subsection{Outliers}
Outliers in a Gaussian mixture model can be handled by introducing a latent variable $u$ to scale each mixture component covariance matrix $\boldsymbol{\Sigma}_k$. That family of mixture models is known as  scale mixtures of Normal distributions \cite{Andrews1974}. Introducing the latent positive variable $u$ into (\ref{latent2}), the following scale component distribution is obtained 

			\begin{equation}\label{outlier}
				p(\boldsymbol{x}|u,z,\boldsymbol{\Theta},\mathcal{K})=\prod_{k\in\mathcal{K}}\mathcal{N}(\boldsymbol{x}|\boldsymbol{\mu}_k,u^{-1}\boldsymbol{\Sigma}_k)^{\delta_{z}^k}\;,
			\end{equation}
			
			\begin{flushleft}
				and the joint distribution of $(\boldsymbol{x},u)$ is derived from (\ref{outlier}) such that 
			\end{flushleft}
			
			\begin{equation}\label{outlier2}
				p(\boldsymbol{x},u|z,\boldsymbol{\Theta},\mathcal{K})=\prod_{k\in\mathcal{K}} \left[\mathcal{N}(\boldsymbol{\mu}_k,u^{-1}\boldsymbol{\Sigma}_k)p_k(u)\right]^{\delta_z^k}
			\end{equation}
			
			\begin{flushleft}
				where $p_k(u)$ is the prior distribution of $u$ conditionally to $z=k$.
			\end{flushleft}
			
			 For the sake of keeping conjugacy between  prior and posterior distributions of $u$, a Gamma distribution $\mathcal{G}(u|\alpha_k,\beta_k)$ with shape and rate parameters $(\alpha_k,\beta_k)$ is chosen for $p_k(u)$. Integrating (\ref{outlier2}) out $u$, the resulting marginal  $p(\boldsymbol{x}|z,\boldsymbol{\Theta},\mathcal{K})$ is a heavy-tailed distribution known as the Student-t distribution \cite{Dumitru2017}.


\subsection{Proposed model}

Combining (\ref{latent2}), (\ref{missing}) and (\ref{outlier}), the following joint latent representation is obtained 

\begin{equation*}
			\begin{split}
			p(\boldsymbol{x}^{obs},\mathbf{h}|\boldsymbol{\Theta},\mathcal{K}))&=\prod_{k\in\mathcal{K}} \bigg[a_k\mathcal{N}\left(\boldsymbol{x}^{miss}|\boldsymbol{\epsilon}_k^{\text{miss}},u^{-1}\boldsymbol{\Delta}_k^{\text{miss}}\right)\\
			&\mathcal{N}\left(\boldsymbol{x}^{\text{obs}}|\boldsymbol{\mu}_k^{\text{obs}},u^{-1}\boldsymbol{\Delta}_k^{\text{obs}}\right)\mathcal{G}(u|\alpha_k,\beta_k)\bigg]^{\delta_z^k}
			\end{split}
\end{equation*}

\begin{flushleft}
where $\boldsymbol{h}=(\boldsymbol{x}^{miss},u,z)$ are the latent variables.
\end{flushleft}

Finally, assuming a dataset $\boldsymbol{X}\in \mathbb{R}^{d\times J}$ of i.i.d observations $(\boldsymbol{x}_1,\ldots,\boldsymbol{x}_J)$ and independent latent data $\mathbf{H}=(\mathbf{X}^{\text{miss}},\boldsymbol{u},\boldsymbol{z})$, the complete likelihood function can be expressed as 

\begin{equation*}
			\begin{split}
			p(\mathbf{X}^{obs},\mathbf{H}|\boldsymbol{\Theta},\mathcal{K}))&=\prod_{j\in\mathcal{J}}\prod_{k\in\mathcal{K}} \bigg[a_k\mathcal{N}\left(\boldsymbol{x}_j^{miss}|\boldsymbol{\epsilon}_k^{\text{miss}},u_j^{-1}\boldsymbol{\Delta}_k^{\text{miss}}\right)\\
			&\mathcal{N}\left(\boldsymbol{x}_j^{\text{obs}}|\boldsymbol{\mu}_k^{\text{obs}},u_j^{-1}\boldsymbol{\Delta}_k^{\text{obs}}\right)\mathcal{G}(u_j|\alpha_k,\beta_k)\bigg]^{\delta_{z_j}^k}
			\end{split}
\end{equation*}

\begin{flushleft}
where $\mathcal{J}=\{1,\ldots,J\}$, $\mathbf{X}^{\text{obs}}=\left\{\boldsymbol{x}_1^{\text{obs}},\ldots,\boldsymbol{x}_j^{\text{obs}}\right\}$, $\mathbf{X}^{\text{miss}}=\left\{\boldsymbol{x}_1^{\text{miss}},\ldots,\boldsymbol{x}_j^{\text{miss}}\right\}$, $\boldsymbol{z}=\{ z_j\}_{j\in\mathcal{J}}$ is the discrete variable introduced to indicate  which cluster the data $\boldsymbol{x}_j$ belongs to and $\boldsymbol{u}=\{ u_j\}_{j\in\mathcal{J}}$ is the scale variable associated to $\boldsymbol{x}_j$. 
\end{flushleft}

At last, the Bayesian framework imposes to specify a prior distribution for the parameters $\boldsymbol{\Theta}=(\boldsymbol{a},\boldsymbol{\alpha},\boldsymbol{\beta},\boldsymbol{\mu},\boldsymbol{\Sigma})$. The resulting conjugate priors are

  \begin{equation}\label{thetaprior}
          \left \{
          \begin{aligned}
          	&p(\boldsymbol{a}|\mathcal{K})=\mathcal{D}(\boldsymbol{a}|\boldsymbol{k}_0)~, \\
		&p(\boldsymbol{\alpha},\boldsymbol{\beta}|\mathcal{K})=\prod_{k\in\mathcal{K}}p(\alpha_k,\beta_k|p_0,q_0,s_0,r_0)~,\\
          	&p(\boldsymbol{\mu},\boldsymbol{\Sigma}|\mathcal{K}) =\prod_{k\in\mathcal{K}} \mathcal{N}(\boldsymbol{\mu}_k|\boldsymbol{\mu}_0,\eta_0^{-1}\Sigma_k)\mathcal{IW}(\boldsymbol{\Sigma}_k|\gamma_0,\boldsymbol{\Sigma}_0)~.
          \end{aligned}
          \right .
      \end{equation}
      
      \begin{flushleft}
         where $\boldsymbol{a}$ follows a Dirichlet distribution, $(\boldsymbol{\mu}_k,\boldsymbol{\Sigma}_k)$  a Normal-Inverse-Wishart distribution and $p(\cdot|p_0,q_0,s_0,r_0)$ are defined below,
         \end{flushleft}
         
         \begin{align*}
         &\mathcal{D}(\mathbf{a}|\boldsymbol{\kappa})=c_{\mathcal{D}}(\boldsymbol{\kappa})\prod\limits_{k\in\mathcal{K}}a_k^{\kappa_k-1}~,\\
            &\mathcal{IW}(\boldsymbol{\Sigma}|\gamma,\mathbf{S})=c_{\mathcal{IW}}(\gamma,\mathbf{S})|\boldsymbol{\Sigma}|^{-\frac{\gamma+d+1}{2}}\exp\left(-\frac{1}{2}tr(\mathbf{S}\boldsymbol{\Sigma}^{-1})\right)\;,
         \end{align*}
         
         \begin{flushleft}
         where  $c_{\mathcal{D}}(\boldsymbol{\kappa})$ and $c_{\mathcal{IW}}(\gamma,\mathbf{S})$ are normalizing constants such that
         \end{flushleft}
                  
         \begin{equation*}
         c_{\mathcal{D}}(\boldsymbol{\kappa})=\frac{\Gamma\left(\sum_{k\in\mathcal{K}}\kappa_k\right)}{\prod_{k\in\mathcal{K}}\Gamma(\kappa_k)},~c_{\mathcal{IW}}(\gamma,\mathbf{S})=\frac{|\mathbf{S}|^{\frac{\gamma}{2}}}{2^{\frac{d\gamma}{2}}\Gamma_d(\frac{\gamma}{2})}~.
         \end{equation*}
         
			To avoid a non closed-form posterior distribution for  $(\boldsymbol{\alpha},\boldsymbol{\beta})$, the following conditional prior is  introduced :
    
    \begin{equation}\label{condprior}
 	p(\alpha,\beta|p_0,q_0,s_0,r_0)=p(\beta|\alpha,s_0,q_0)p(\alpha|p_0,q_0,s_0,r_0)
    \end{equation}
    
    \begin{flushleft}
    where $p_0,q_0,s_0,r_0>0$ and
    \end{flushleft}
    
    \begin{align*}
    	&p(\beta|\alpha,s_0,q_0)=\mathcal{G}(\beta|s_0\alpha + 1,q_0)\\
        &p(\alpha|p_0,q_0,s_0,r_0)=\frac{1}{M_0}\frac{p_0^{\alpha-1}\Gamma(s_0\alpha+1)}{q_0^{s_0\alpha+1}\Gamma(\alpha)^{r_0}}\mathbb{I}_{\{\alpha>0\}}
    \end{align*}
    
        \begin{flushleft}
    where
    \end{flushleft}
    
        \begin{equation*}
    	M_0=\int\frac{p_0^{\alpha-1}\Gamma(s_0\alpha+1)}{q_0^{s_0\alpha+1}\Gamma(\alpha)^{r_0}}\mathbb{I}_{\{\alpha>0\}}d\alpha
    \end{equation*}
    
\begin{figure}
\centering
		\begin{tikzpicture}[scale=0.6]
			\draw (-1,-1.25) rectangle (7,0.75);
			\draw (-1,-4.5) rectangle (6,-1.5);
			\draw (6.6,-0.9) node{$K$};
			\draw (-0.6,-4.2) node{$J$};
			\draw (-0.5,1) rectangle (0.5,3);
			\draw[->](0,1)--(0,0.25);
			\draw (0,2.5) node{$\gamma_0$};
			\draw (0,1.5) node{$\boldsymbol{\Sigma}_0$};
			\draw (1.5,1) rectangle (2.5,3);
			\draw[->](2,1)--(2,0.25);
			\draw (2,1.5) node{$\boldsymbol{\mu}_0$};
			\draw (2,2.5) node{$\eta_0$};
			\draw (0,-0.25) circle (0.5) ;
			\draw (0,-0.25) node{$\boldsymbol{\Sigma}_k$};
			\draw [->] (0.5,-0.25) -- (1.5,-0.25);
			\draw [->] (0,-0.75) -- (0.75,-2.25);
			\draw (2,-0.25) circle (0.5) ;
			\draw (2,-0.25) node{$\boldsymbol{\mu}_k$};
			\draw [->] (2,-0.75) -- (1.25,-2.25);
			\draw (4,-0.25) circle (0.5) ;
			\draw (4,-0.25) node{$\alpha_{k}$};
			\draw [->] (4.5,-0.25) -- (5.5,-0.25);
			\draw [->] (4,-0.75) -- (4.9,-1.75);
			\draw (6,-0.25) circle (0.5) ;
			\draw (6,-0.25) node{$\beta_{k}$};
			\draw [->] (6,-0.75) -- (5.1,-1.75);
			\draw (5,-2.25) circle (0.5) ;
			\draw (5,-2.25) node{$u_{j}$};
			\draw [->] (4.5,-2.25) -- (2.75,-2.75);
			\draw[dashed,rounded corners] (-0.75,-3.75) rectangle (2.75,-2.25);
			\draw (0,-3) node{$\boldsymbol{x}^\text{miss}_{j}$};
			\draw (0,-3) circle (0.7) ;
			\draw (2,-3) node{$\boldsymbol{x}^\text{obs}_{j}$};
			\draw (2,-3) circle (0.7) ;
			\draw [->] (1.3,-3) -- (0.7,-3);
			\draw (5,-3.75) node{$z_{j}$};
			\draw (5,-3.75) circle (0.5) ;
			\draw [->] (4.5,-3.75) -- (2.75,-3.25);
			\draw [->] (6.5,-3.75) -- (5.5,-3.75);
			\draw [->] (5,-3.25) -- (5,-2.75);
			\draw (7,-3.75) circle (0.5) ;
			\draw (7,-3.75) node{$\mathbf{a}$};
			\draw (7,-2.25) circle (0.5) ;
			\draw (7,-2.25) node{$\kappa_{0}$};
			\draw [->] (7,-2.75) -- (7,-3.25);
			\draw (3.5,1) rectangle (4.5,3);
			\draw (4,2.5) node{$p_0$};
			\draw (4,1.5) node{$r_0$};
			\draw [->] (4,1) -- (4,0.25);
			\draw (5.5,1) rectangle (6.5,3);
			\draw (6,2.5) node{$q_0$};
			\draw (6,1.5) node{$s_0$};
			\draw [->] (6,1) -- (6,0.25);
			\draw [->] (5.5,1) -- (4.2,0.2);
	\end{tikzpicture}
	
\caption{ Graphical representation of the proposed model. The arrows represent conditional dependencies between the random variables. The K-plate represents the K mixture components and the J-plate the independent identically distributed observations $\boldsymbol{x}_j$, the scale variables $u_{j}$ and the indicator variables $z_{j}$.}
\label{graphical}
\end{figure}
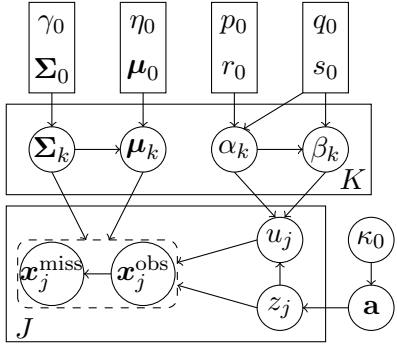

    The normalization constant $M_0$ is intractable and a Laplace approximation method \cite{Tierney1986} is derived to estimate it. Figure \ref{graphical} shows the proposed model.

\section{Inference}
		\label{section:3}
		
		The Variational Bayesian inference  was introduced in \cite{Waterhouse1996} as  an ensemble  learning method for the mixtures of experts in order to avoid over-fitting and noise level under-estimation problems of traditional maximum likelihood inference. In \cite{Attias1999}, the Variational Bayesian inference was generalized for different types of mixture distributions and took the name Variational Bayes (VB).
		VB can be viewed as a Bayesian generalization of the Expectation-Maximization (EM) algorithm \cite{Dempster1977} combined with a Mean Field Approach \cite{Opper2001}. It consists in approximating the intractable posterior distribution $P=p(\boldsymbol{H},\boldsymbol{\Theta}|\boldsymbol{X},K)$ by a tractable one $Q=q(\boldsymbol{H},\boldsymbol{\Theta})$ whose parameters are chosen via a variational principle to minimize the Kullback-Leibler (KL) divergence 

\begin{equation*}
	KL\left[Q||P\right]=\int q(\boldsymbol{H},\boldsymbol{\Theta}) \log\left(\frac{q(\boldsymbol{H},\boldsymbol{\Theta})}{p(\boldsymbol{H},\boldsymbol{\Theta}|\boldsymbol{X},\mathcal{K})}\right)d\boldsymbol{H}d\boldsymbol{\Theta}~,
\end{equation*}

\begin{flushleft}
	Noting that $p(\boldsymbol{H},\boldsymbol{\Theta}|\boldsymbol{X},\mathcal{K})=\frac{p(\boldsymbol{X},\boldsymbol{H},\boldsymbol{\Theta}|\mathcal{K})}{p(\boldsymbol{X}|\mathcal{K})}$, the KL divergence can be written as 
\end{flushleft}

\begin{equation*}
	KL\left[Q||P\right]=\log p(\boldsymbol{X}|\mathcal{K})-\mathcal{L}(Q)~.
\end{equation*}

$\mathcal{L}(Q)$ is considered as a lower bound for  the log evidence $\log p(\boldsymbol{X}|\mathcal{K})$ and can be expressed as

\begin{equation}\label{bound}
	\mathcal{L}(Q)=\mathbb{E}_q\left[\log p(\boldsymbol{X},\boldsymbol{H},\boldsymbol{\Theta}|\mathcal{K})\right]-\mathbb{E}_q\left[\log q(\boldsymbol{H},\boldsymbol{\Theta})\right]\;,
\end{equation}

\begin{flushleft}
	where $\mathbb{E}_q[\cdot]$ denotes the expectation with respect to $q$.
\end{flushleft}

Then, minimizing the KL divergence is equivalent to maximizing  $\mathcal{L}(Q)$. Assuming that $q(\mathbf{\boldsymbol{H},\boldsymbol{\Theta}})$ can be factorized over the latent variables $\boldsymbol{H}$ and the  parameters $(\boldsymbol{\Theta})$, a free-form maximization  with respect to  $q(\boldsymbol{H})$ and $q(\boldsymbol{\Theta})$ leads to  the following update rules  :

\begin{align*}
&\textbf{VBE-step} : q(\boldsymbol{H})\propto \exp\left(\mathbb{E}_{\boldsymbol{\Theta}}\left[\log p(\mathbf{X},\boldsymbol{H}|\boldsymbol{\Theta},\mathcal{K})\right]\right),\\
&\textbf{VBM-step} :  q(\boldsymbol{\Theta})\propto \exp\left(\mathbb{E}_{\boldsymbol{H}}\left[\log p(\boldsymbol{\Theta},\mathbf{X},\boldsymbol{H}|\mathcal{K})\right]\right)~.
\end{align*}

The expectations  $\mathbb{E}_{\boldsymbol{H}}[\cdot]$ and $\mathbb{E}_{\boldsymbol{\Theta}}[\cdot]$ are respectively taken with respect to the variational  posteriors $q(\boldsymbol{H})$ and $q(\boldsymbol{\Theta})$. Thereafter, the algorithm  iteratively updates the variational posteriors  by increasing  the bound  $\mathcal{L}(Q)$. Running the algorithm steps,  each posterior distribution is obtained in the following  subsections.

Noting that $p(\mathbf{X},\mathbf{z},\mathbf{u}|\boldsymbol{\Theta},\mathcal{K})$ can be factorized as $p(\mathbf{X}|\mathbf{u},\mathbf{z},\boldsymbol{\Theta},\mathcal{K})p(\mathbf{u}|\mathbf{z},\boldsymbol{\Theta},\mathcal{K})p(\mathbf{z}|\boldsymbol{\Theta},\mathcal{K})$, a factorized form $q(\mathbf{X}^{miss}|\mathbf{u},\mathbf{z}))q(\mathbf{u}|\mathbf{z})q(\mathbf{z})$ is similarly chosen for $q(\mathbf{X}^{miss},\mathbf{u},\mathbf{z})$. Moreover, $p(\boldsymbol{\Theta}|\mathbf{X},\mathbf{z},\mathbf{u},K)$ can be decomposed as $p(\boldsymbol{a}|\mathbf{X},\mathbf{z},\mathbf{u},K)p(\boldsymbol{\mu}|\boldsymbol{\Sigma},\mathbf{X},\mathbf{z},\mathbf{u},K)p(\boldsymbol{\Sigma}|\mathbf{X},\mathbf{z},\mathbf{u},K)$, the following similar form is chosen for $q(\boldsymbol{\Theta})$

         \begin{equation*}
         q(\boldsymbol{\Theta})=q(\boldsymbol{a})\prod_{k\in\mathcal{K}}q(\alpha_k,\beta_k)q(\boldsymbol{\mu}_k|\boldsymbol{\Sigma}_k)q(\boldsymbol{\Sigma}_k)~.
          \end{equation*}
          
         Since conjugate priors have been designed in (\ref{thetaprior}) and (\ref{condprior}), conjugate posterior distributions are obtained from the VBM-step for $\boldsymbol{\Theta}$ : 

           \begin{equation}\label{posterior_parameters}
          \left \{
          \begin{aligned}
          	&q(\boldsymbol{a})=\mathcal{D}(\boldsymbol{a}|\tilde{\boldsymbol{k}})~, \\
		&q(\alpha_k,\beta_k)=p(\alpha_k,\beta_k|\tilde{p}_k,\tilde{q}_k,\tilde{s}_k,\tilde{r}_k)~,\\
          	&q(\boldsymbol{\mu}_k|\boldsymbol{\Sigma}_k) = \mathcal{N}(\boldsymbol{\mu}_k|\tilde{\boldsymbol{\mu}}_k,\tilde{\eta}_k^{-1}\boldsymbol{\Sigma}_k)~, \\
          &q(\boldsymbol{\Sigma}_k) = \mathcal{IW}(\boldsymbol{\Sigma}_k|\tilde{\gamma}_k,\tilde{\boldsymbol{\Sigma}}_k)~.
          \end{aligned}
          \right .
      \end{equation}

 			Calculations of variational posterior  distributions related to latent variables $\mathbf{H}=(\mathbf{X}^{\text{miss}},\mathbf{H})$ and parameters $\boldsymbol{\Theta}$ are presented below.
			
 \subsection{Variational posterior distributions for latent variables}

The VBE-step can be computed by developing the expectation 

\begin{equation}\label{expectation}
 					\mathbb{E}_{\boldsymbol{\Theta}}\left[\log p(\mathbf{X},\mathbf{H}|\boldsymbol{\Theta},K)\right]=\sum_{j\in\mathcal{J}}\sum_{k\in\mathcal{K}} \delta_{z_j}^kf_k(\boldsymbol{x}_j,u_j)\;.
			 \end{equation}
			 
			 \begin{flushleft}
			 	where
			 \end{flushleft}
			 
			 \begin{equation}\label{function_f}
			 	\begin{split}
					f_k(\boldsymbol{x}_j,u_j)=& -\frac{\mathbb{E}_{\boldsymbol{\Theta}}[\log \boldsymbol{|\Sigma}_k|]}{2}-\frac{d}{2}(\log 2\pi-\log u_{j}) + \mathbb{E}_{\boldsymbol{\Theta}}[\log a_k]
                    -\frac{u_{j}}{2}\mathbb{E}_{\boldsymbol{\Theta}}\left[\mathcal{D}(\boldsymbol{x}_j,\boldsymbol{\mu}_k,\boldsymbol{\Sigma}_k)\right]\\ 
                    &+ \mathbb{E}_{\boldsymbol{\Theta}}[\alpha_k]\mathbb{E}_{\boldsymbol{\Theta}}[\log \beta_k]-\mathbb{E}_{\boldsymbol{\Theta}}[\log \Gamma (\alpha_k)]+ \left(\mathbb{E}_{\boldsymbol{\Theta}}[\alpha_k]-1\right)\log u_{j}-\mathbb{E}_{\boldsymbol{\Theta}}[\beta_{k}]u_{j}\; .
                    		\end{split}
			 \end{equation}
			 
			Conditionally to $z_j=k$ and a given $u_j$, $\boldsymbol{x}_j$ follows a Gaussian distribution with mean $\tilde{\boldsymbol{\mu}}_k$ and covariance matrix $\tilde{\gamma}_k^{-1}u_j^{-1}\tilde{\boldsymbol{\Sigma}}_k$   since
			
			\begin{equation}
				\mathbb{E}_{\boldsymbol{\Theta}}[\mathcal{D}(\boldsymbol{x}_j,\boldsymbol{\mu}_k,\boldsymbol{\Sigma}_k)] =\tilde{\gamma}_k(\boldsymbol{x}_j-\tilde{\boldsymbol{\mu}}_k)^T\tilde{\boldsymbol{\Sigma}}_k^{-1}(\boldsymbol{x}_j-\tilde{\boldsymbol{\mu}}_k)+\frac{d}{\tilde{\eta}_k}\; .
			\end{equation}
			
			 Therefore the following distribution for $\boldsymbol{x}_j^{\text{miss}}$ is deduced from (\ref{missing}), (\ref{missing2}) and (\ref{missing3}) 
			 
			 \begin{equation}
			q(\boldsymbol{x}_j^{\text{miss}}|u_j,z=k)= \mathcal{N}\left(\tilde{\boldsymbol{\epsilon}}_{jk}^{\text{miss}},u_j^{-1}\tilde{\boldsymbol{\Delta}}_k^{\text{miss}}\right)
		\end{equation}
		
		\begin{flushleft}
			where
		\end{flushleft}
		
		\begin{align}
			&\tilde{\boldsymbol{\epsilon}}_{jk}^{\text{miss}}=\tilde{\boldsymbol{\mu}}_k^{\text{miss}}+\tilde{\boldsymbol{\Sigma}}_k^{\text{cov}}\tilde{\boldsymbol{\Sigma}}_k^{\text{obs}-1}(\boldsymbol{x}_j^{\text{obs}}-\tilde{\boldsymbol{\mu}}_k^{\text{obs}} )\\
			&\tilde{\boldsymbol{\Delta}}_k^{\text{miss}}= \frac{\tilde{\boldsymbol{\Sigma}}_k^{\text{miss}}-\tilde{\boldsymbol{\Sigma}}_k^{\text{cov}}\tilde{\boldsymbol{\Sigma}}_k^{\text{obs}-1}\tilde{\boldsymbol{\Sigma}}_k^{\text{cov}'}}{\tilde{\gamma}_k}\;.
		\end{align}
		
		Auxiliary variables $\tilde{\boldsymbol{x}}_j\in\mathbb{R}^d$ and $\boldsymbol{\Delta}_k^{x_j}\in \mathbb{R}^{d\times d}$ are introduced for the VBM - step such that 
		
		\begin{align*}
			&\tilde{\boldsymbol{x}}_j=\mathbb{E}_{\mathbf{H}}[\boldsymbol{x}_j]=\begin{pmatrix} 
			\tilde{\boldsymbol{\epsilon}}_{jk}^{\text{miss}}\\
			\boldsymbol{x}_j^{\text{obs}} 
		\end{pmatrix}\\
			&\boldsymbol{\Delta}_k^{x_j}=\mathbb{E}_{\mathbf{H}}[\boldsymbol{x}_j\boldsymbol{x}_j^T]-\tilde{\boldsymbol{x}}_j\tilde{\boldsymbol{x}}_j^T=\begin{pmatrix} 
			\tilde{\boldsymbol{\Delta}}_k^{\text{miss}}&\mathbf{0}^{d_{\text{miss}}^j\times d_{\text{obs}}^j}\\
			\mathbf{0}^{d_{\text{obs}}^j\times d_{\text{miss}}^j}&\mathbf{0}^{d_{\text{obs}}^j\times d_{\text{obs}}^j}
			\end{pmatrix}\;.
		\end{align*}
		
		Marginalising over $\boldsymbol{x}_j^{\text{miss}}$, (\ref{function_f})  becomes 
		
		\begin{equation}\label{function_fuz}
			\begin{split}
				\log \int \exp f_k(\boldsymbol{x}_j,u_j)d\boldsymbol{x}_j^{\text{miss}}=& -\frac{\mathbb{E}_{\boldsymbol{\Theta}}[\log \boldsymbol{|\Sigma}_k|]}{2}-\frac{d_{\text{obs}}^j}{2}(\log 2\pi-\log u_{j}) + \mathbb{E}_{\boldsymbol{\Theta}}[\log a_k]\\
                    &-\frac{u_{j}}{2}\left(\mathcal{D}\left(\boldsymbol{x}^{\text{obs}}_j,\tilde{\boldsymbol{\mu}}_k^{\text{obs}},\tilde{\boldsymbol{\Delta}}_k^{\text{obs}}\right)+\frac{d}{\tilde{\eta}_k}\right) 
                    +\frac{\log |\tilde{\boldsymbol{\Delta}}_k^{\text{miss}}| }{2}+ \mathbb{E}_{\boldsymbol{\Theta}}[\alpha_k]\mathbb{E}_{\boldsymbol{\Theta}}[\log \beta_k]\\&-\mathbb{E}_{\boldsymbol{\Theta}}[\log \Gamma (\alpha_k)]+ \left(\mathbb{E}_{\boldsymbol{\Theta}}[\alpha_k]-1\right)\log u_{j}-\mathbb{E}_{\boldsymbol{\Theta}}[\beta_{k}]u_{j}\; ,
			\end{split}
		\end{equation}
		
		\begin{flushleft}
			where $d_{\text{obs}}^j$ is the dimension of $\boldsymbol{x}^{\text{obs}}_j$ and 
		\end{flushleft}
		
			\begin{equation*}
			\tilde{\boldsymbol{\Delta}}_k^{\text{obs}}= \frac{\left(\tilde{\boldsymbol{\Sigma}}_k^{\text{obs}-1}+2\times\tilde{\boldsymbol{\Sigma}}_k^{\text{obs}-1}\tilde{\boldsymbol{\Sigma}}_k^{\text{cov}'}\left(\tilde{\boldsymbol{\Delta}}_k^{\text{miss}}\right)^{-1}\tilde{\boldsymbol{\Sigma}}_k^{\text{cov}}\tilde{\boldsymbol{\Sigma}}_k^{\text{obs}-1}\right)^{-1}}{\tilde{\gamma}_k}\;.
		\end{equation*}
		
A conditional posterior  for $u_j$ is deduced from (\ref{function_fuz}) such that

 			\begin{equation*}
			  q(u_{j}|z_{j}=k)\sim 	\mathcal{G}(\tilde{\alpha}_{jk},\tilde{\beta}_{jk})
			 \end{equation*}
			  
			  \begin{flushleft}
			  where
			  \end{flushleft}

			  \begin{align*}
			  	&\tilde{\alpha}_{jk}=\mathbb{E}_{\boldsymbol{\Theta}}[\alpha_k]+\frac{d_{\text{obs}}^j}{2}\;,\\
				&\tilde{\beta}_{jk}=\frac{1}{2}\left(\mathcal{D}\left(\boldsymbol{x}^{\text{obs}}_j,\tilde{\boldsymbol{\mu}}_k^{\text{obs}},\tilde{\boldsymbol{\Delta}}_k^{\text{obs}}\right)+\frac{d_{\text{obs}}^j}{\tilde{\eta}_k}\right)+\mathbb{E}_{\boldsymbol{\Theta}}[\beta_k] \;.
			  \end{align*}
 Expectations of $\mathbf{u}|\mathbf{z}$ are derived from the Gamma distribution properties such that 
 \begin{align*}
&\mathbb{E}_{\mathbf{H}}[u_{j}]=\frac{\tilde{\alpha}_{jk}}{\tilde{\beta}_{jk}}~,\\
 &\mathbb{E}_{\mathbf{H}}[\log u_{j}]=\psi(\tilde{\alpha}_{jk})-\log \tilde{\beta}_{jk}~,
 \end{align*}
\begin{flushleft}
where $\psi(\cdot)$ is the digamma function.
\end{flushleft}
Due to the conjugacy property, a conjugate posterior distribution for latent variable $\mathbf{z}$ is obtained from  (\ref{expectation})

\begin{equation}\label{qz}
q(\mathbf{z})=\prod_{j\in\mathcal{J}}\prod_{k\in\mathcal{K}}\rho_{jk}^{\delta^k_{z_j}}\;,
\end{equation}

\begin{flushleft}
where $\rho_{jk}=q(z_{j}=k)$ is called the responsibility.
\end{flushleft}

Instead of assuming that most of the probability mass of the posterior distribution of  the scale variable $\mathbf{u}$ is located around its mean \cite{Svensen2005,Sun2017}, \cite{Archambeau2007} proposed to integrate out $\mathbf{u}$ the joint  variational posterior $q(\mathbf{u},\mathbf{z})$ to obtain $q(\mathbf{z})$. Therefore, it consists in substituting (\ref{function_fuz}) in the E-step and marginalizing  over $\mathbf{u}$.  That approach leads to the following responsibilities 

\begin{equation*}
	\begin{split}
		\rho_{jk} &\propto \int \exp f_k(\boldsymbol{x}_j,u_j)d\boldsymbol{x}_j^{\text{miss}}du_{j}\\
        &\propto \frac{\exp \left(\mathbb{E}_{\boldsymbol{\Theta}}[\log a_k]+\mathbb{E}_{\boldsymbol{\Theta}}[\log \beta_k]\mathbb{E}_{\boldsymbol{\Theta}}[\alpha_k] \right)\Gamma\left(\mathbb{E}_{\boldsymbol{\Theta}}[\alpha_k]+\frac{d^j_{\text{obs}}}{2}\right)}{\exp \left(\frac{\mathbb{E}_{\boldsymbol{\Theta}}[\log|\boldsymbol{\Sigma}_k|]-\log |\tilde{\boldsymbol{\Delta}}_k^{\text{miss}}|}{2}+\mathbb{E}_{\boldsymbol{\Theta}}[\log\Gamma(\alpha_k)]\right) \mathbb{E}_{\boldsymbol{\Theta}}[\beta_k]^{\left(\mathbb{E}_{\boldsymbol{\Theta}}[\alpha_k]+\frac{d^j_{\text{obs}}}{2}\right)}}\\
        &\times \left[1+\frac{\mathcal{D}\left(\boldsymbol{x}^{\text{obs}}_j,\tilde{\boldsymbol{\mu}}_k^{\text{obs}},\tilde{\boldsymbol{\Delta}}_k^{\text{obs}}\right)+\frac{d_{\text{obs}}^j}{\tilde{\eta}_k}}{2\mathbb{E}_{\boldsymbol{\Theta}}[\beta_k]}\right]^{-\left(\mathbb{E}_{\boldsymbol{\Theta}}[\alpha_k]+\frac{d^j_{\text{obs}}}{2}\right)} \; .
	\end{split}
\end{equation*}


Then, the responsibilities are normalized as follows 

\begin{equation}\label{norm}
	r_{jk}=\frac{\rho_{jk}}{\sum_{k=1}^K\rho_{jk}}\;.
\end{equation}

Expectation of $\mathbf{z}$ is deduced from (\ref{qz}) and is given by 
\begin{equation*}
\mathbb{E}_{\mathbf{H}}[\delta^k_{z_j}]=r_{jk}\; .
\end{equation*}
			
 \subsection{Variational posterior distributions for parameters}
	      The VBM-step can be computed by developing the expectation 

			\begin{equation*}
 					\mathbb{E}_{\mathbf{H}}\left[\log p(\boldsymbol{\Theta},\mathbf{X},\mathbf{H}|K)\right]=\sum_{j\in\mathcal{J}}\sum_{k\in\mathcal{K}}\mathbb{E}_{\mathbf{H}}\left[ \delta_{z_j}^k\log \left(a_k\mathcal{N}(\boldsymbol{x}_j|\boldsymbol{\mu}_k,\boldsymbol{\Sigma}_k)\mathcal{G}\left(u_{j}|\alpha_k,\beta_k\right)\right)\right]+\log p(\boldsymbol{\Theta})\;.
			 \end{equation*}

      Update rules for hyper-parameters defined in (\ref{posterior_parameters}) are 

 	\begin{align*}
 					&\tilde{k}_k=k_0+N\bar{\pi}_k~, \\
					&\tilde{\eta}_k=\eta_0+N\bar{\omega}_k~, \\
					&\tilde{p}_k=p_0\exp \left(N\bar{\delta}_k\right)~,\\
      					 &\tilde{q}_k=q_0+N\bar{\omega}_k~,\\
    					&\tilde{r}_k=r_0+N\bar{\pi}_k~,\\
        					&\tilde{s}_k=s_0+N\bar{\pi}_k~,\\
					&\tilde{\boldsymbol{\mu}}_k=\frac{\eta_0\boldsymbol{\mu}_0+N\bar{\omega}_k\boldsymbol{\mu}^x_k}{\tilde{\eta}_k}~, \\
					&\tilde{\gamma}_k=\gamma_0+N\bar{\pi}_k~, \\
 					&\tilde{\boldsymbol{\Sigma}}_k=\boldsymbol{\Sigma}_0+\frac{N\bar{\omega}_k\eta_0}{\tilde{\eta}_k}\left(\boldsymbol{\mu}^x_k-\boldsymbol{\mu}_0\right)\left(\boldsymbol{\mu}^x_k-\boldsymbol{\mu}_0\right)^T+\boldsymbol{\Sigma}^x_{k}+\boldsymbol{\Sigma}^{\text{m}}_{k}\; ,
 	\end{align*}
	\begin{flushleft}
	where  auxiliary variables are obtained as follows
	\end{flushleft}			
	\begin{align*}
				&\bar{\pi}_k=\frac{1}{J}\sum\limits_{j\in\mathcal{J}}\mathbb{E}_{\mathbf{H}}[\delta^k_{z_j}]~,\\
				&\bar{\omega}_k=\frac{1}{J}\sum_{j\in\mathcal{J}}\mathbb{E}_{\mathbf{H}}[\delta^k_{z_j}]\mathbb{E}_{\mathbf{H}}[u_{j}]~,\\
				 &\bar{\delta}_k=\frac{1}{J}\sum_{j\in\mathcal{J}}\mathbb{E}_{\mathbf{H}}[\delta^k_{z_j}]\mathbb{E}_{\mathbf{H}}[\log u_{jk}]~.\\
				 &\boldsymbol{\mu}^x_k=\frac{1}{N\bar{\omega}_k}\sum_{j\in\mathcal{J}}\mathbb{E}_{\mathbf{H}}[\delta^k_{z_j}]\mathbb{E}_{\mathbf{H}}[u_{j}]\tilde{\boldsymbol{x}}_j~,\\
				 &\boldsymbol{\Sigma}^x_k=\frac{1}{N\bar{\omega}_k}\sum_{j\in\mathcal{J}}\mathbb{E}_{\mathbf{H}}[\delta^k_{z_j}]\mathbb{E}_{\mathbf{H}}[u_{j}]\left(\tilde{\boldsymbol{x}}_j-\boldsymbol{\mu}^x_k\right)\left(\tilde{\boldsymbol{x}}_j-\boldsymbol{\mu}^x_k\right)^T~,\\
				 &\boldsymbol{\Sigma}^{\text{m}}_{k}=\sum_{j\in\mathcal{J}}\mathbb{E}_{\mathbf{H}}[\delta^k_{z_j}]\boldsymbol{\Delta}_k^{x_j}\;.
	\end{align*}
                
Using the properties of the Dirichlet and the Inverse Wishart distribution, the following expectations are defined 
 \begin{align*}
 &\mathbb{E}_{\boldsymbol{\Theta}}[\log a_k]=\psi(\tilde{\kappa}_k)-\psi\left(\sum_{k'=1}^K\tilde{\kappa}_{k'}\right)~,\\
 &\mathbb{E}_{\boldsymbol{\Theta}}[\boldsymbol{\Sigma}_k^{-1}]=\tilde{\gamma}_k\tilde{\boldsymbol{\Sigma}}_k^{-1}~,\\
 &\mathbb{E}_{\boldsymbol{\Theta}}[\log \boldsymbol{|\Sigma}_k|]=\log|\tilde{\boldsymbol{\Sigma}}_k|-\sum_{i=1}^d\psi \left(\frac{\tilde{\gamma}_k+1-i}{2}\right)-d\log2\;.
			 \end{align*} 
		
		\subsection{Lower bound elements and expectations}
		
		The lower bound (\ref{bound}) is  proven to increase at each VB iteration and its difference between two iterations can be used as a stop criterion. The introduction of $(\boldsymbol{\alpha},\boldsymbol{\beta})$ slightly modifies the lower bound since the prior distribution (\ref{condprior}) as well as the posterior distributions (\ref{posterior_parameters}) have to be taken into account. Lower bound elements are presented below.\\

    \begin{align*}
    \mathbb{E}_q[\log p(\mathbf{X},\mathbf{H},\boldsymbol{\Theta}|K)]=&\sum_{j\in\mathcal{J}}\sum_{k\in\mathcal{K}}\mathbb{E}_{\mathbf{H}}[\delta^k_{z_j}]\bigg\{\mathbb{E}_{\boldsymbol{\Theta}}[\log a_{k}]-\frac{\mathbb{E}_{\boldsymbol{\Theta}}[\log |\boldsymbol{\Sigma}_k|]}{2}-\frac{d}{2}(\log 2\pi-\mathbb{E}_{\mathbf{H}}[\log u_{jk}])-\mathbb{E}_{\boldsymbol{\Theta}}[\log\Gamma (\alpha_k)]\\
				&-\frac{\mathbb{E}_{\mathbf{H}}[u_{jk}]}{2}\mathbb{E}_{\boldsymbol{\Theta},\mathbf{H}}\left[\mathcal{D}(\boldsymbol{x}_j,\boldsymbol{\mu}_k,\boldsymbol{\Sigma}_k)\right]+\mathbb{E}_{\boldsymbol{\Theta}}[\alpha_k]\mathbb{E}_{\boldsymbol{\Theta}}[\log \beta_k]+(\mathbb{E}_{\boldsymbol{\Theta}}[\alpha_k]-1)\mathbb{E}_{\mathbf{U}}[\log u_{j}]\\
				&-\mathbb{E}_{\boldsymbol{\Theta}}[\beta_k]\mathbb{E}_{\mathbf{U}}[u_{j}]\bigg\} +\sum_{k\in\mathcal{K}}(\kappa_0-1)\mathbb{E}_{\boldsymbol{\Theta}}[\log a_k]-\frac{d}{2}(\log 2\pi-\log \eta_0)-\frac{\mathbb{E}_{\boldsymbol{\Theta}}[\log |\boldsymbol{\Sigma}_k|]}{2}\\
    &-\frac{\eta_0}{2}\mathbb{E}_{\boldsymbol{\Theta}}\left[\mathcal{D}(\boldsymbol{\mu}_k,\boldsymbol{\mu}_0,\boldsymbol{\Sigma}_k)\right]-\frac{tr\{\boldsymbol{\Sigma}_0\mathbb{E}_{\boldsymbol{\Theta}}[\boldsymbol{\Sigma}_k^{-1}]\}}{2}-\frac{\gamma_0+d+1}{2}\mathbb{E}_{\boldsymbol{\Theta}}[\log |\boldsymbol{\Sigma}_k|]\\
    &+(\mathbb{E}_{\boldsymbol{\Theta}}[\alpha_k]-1)\log p_0
                -r_0\mathbb{E}_{\boldsymbol{\Theta}}[\log\Gamma(\alpha_k)]+s_0\mathbb{E}_{\boldsymbol{\Theta}}[\alpha_k]\mathbb{E}_{\boldsymbol{\Theta}}[\log\beta_k]-q_0\mathbb{E}_{\boldsymbol{\Theta}}[\beta_k]\\
                &+K(\log c_{\mathcal{D}}(\kappa_0)+\log c_{\mathcal{IW}}(\gamma_0,\boldsymbol{\Sigma}_0)-\log M_0)\;,
    \end{align*}

\begin{align*}
    \mathbb{E}_q[\log q(\mathbf{H},\boldsymbol{\Theta}|K)]=&\sum\limits_{j\in\mathcal{J}}\sum\limits_{k\in\mathcal{K}}\mathbb{E}_{\mathbf{H}}[\delta^k_{z_{j}}]\left\{\log r_{jk} + \tilde{\alpha}_k\log \tilde{\beta}_k-\log\Gamma (\tilde{\alpha}_k)+(\tilde{\alpha}_k-1)\mathbb{E}_{\mathbf{U}}[\log u_{jk}]-\tilde{\beta}_k\mathbb{E}_{\mathbf{U}}[u_{j}]\right\}\\
    &+\sum\limits_{k\in\mathcal{K}}(\tilde{\kappa}_k-1)\mathbb{E}_{\boldsymbol{\Theta}}[\log a_k]-\frac{d}{2}(\log 2\pi-\log \tilde{\eta}_k)-\frac{\mathbb{E}_{\boldsymbol{\Theta}}[\log |\boldsymbol{\Sigma}_k|]}{2}-\frac{tr\{\tilde{\boldsymbol{\Sigma}}_k\mathbb{E}_{\boldsymbol{\Theta}}[\boldsymbol{\Sigma}_k^{-1}]\}}{2}\\
   & -\frac{\tilde{\gamma}_k+d+1}{2}\mathbb{E}_{\boldsymbol{\Theta}}[\log |\boldsymbol{\Sigma}_k|]-\log M_k+(\mathbb{E}_{\boldsymbol{\Theta}}[\alpha_k]-1)\log \tilde{p}_k-\tilde{r}_k\mathbb{E}_{\boldsymbol{\Theta}}[\log\Gamma(\alpha_k)]\\
   &+\tilde{s}_k\mathbb{E}_{\boldsymbol{\Theta}}[\alpha_k]\mathbb{E}_{\boldsymbol{\Theta}}[\log \beta_{k}]
                -\tilde{q}_k\mathbb{E}_{\boldsymbol{\Theta}}[\beta_{k}] +K\left(\log c_{\mathcal{D}}(\tilde{\boldsymbol{\kappa}})+\log c_{\mathcal{IW}}(\tilde{\gamma_k},\tilde{\boldsymbol{\Sigma}}_k)\right)\;.
    \end{align*}
    
		 Posterior expectations of $\beta_k$ are derived from the posterior Gamma distribution (\ref{posterior_parameters}) properties and can easily be computed conditionally to $\alpha_k$
    
     \begin{align*}
    &\mathbb{E}_{\boldsymbol{\Theta}}[\beta_k]=\frac{\tilde{s}_k\mathbb{E}_{\boldsymbol{\Theta}}[\alpha_k]+1}{\tilde{q}_k}\;,\\
			&\mathbb{E}_{\boldsymbol{\Theta}}[\log \beta_k]=\mathbb{E}_{\boldsymbol{\Theta}}[\psi \left(\tilde{s}_k\alpha_k+1\right)]-\log \tilde{q}_k\;.\\
    \end{align*}
    
    However, expectations depending on $\alpha_k$ are intractable     
    \begin{align}\label{exp1}
    &\mathbb{E}_{\boldsymbol{\Theta}}[\psi \left(\tilde{s}_k\alpha_k+1\right)]=\int\psi \left(\tilde{s}_k\alpha_k+1\right)p(\alpha_k|\tilde{p}_k,\tilde{r}_k)d\alpha_k\;,\\ \label{exp3}
    &\mathbb{E}_{\boldsymbol{\Theta}}[\alpha_k]=\int\alpha_kp(\alpha_k|\tilde{p}_k,\tilde{r}_k)d\alpha_k~,\\ \label{exp2}
   &\mathbb{E}_{\boldsymbol{\Theta}}[\log\Gamma(\alpha_k)]=\int\log\Gamma(\alpha_k)p(\alpha_k|\tilde{p}_k,\tilde{r}_k)d\alpha_k~\; .
    \end{align}
    
    Since lower bound calculation is required as a stop criterion, expectations (\ref{exp1}), (\ref{exp3}) and (\ref{exp2}) have to be approximated. A deterministic method \cite{Tierney1986} based on Laplace approximation is then applied. This method consists in approximating integrals of a smooth function times the posterior $h(\alpha)p(\alpha|p,q,s,r)$ with an approximation proportional to a normal density in $\theta$ such that 
    
    \begin{equation*}
    	\mathbb{E}[h(\alpha]=h(\alpha_0)p(\alpha_0|p,q,s,r)(2\pi)^{d_{\alpha}/2}|-u^{''}(\alpha_0)|^{1/2}\;,
    \end{equation*}
    
    \begin{flushleft}
    where $d_{\alpha}$ is the dimension of $\alpha$, $u(\alpha)=\log \left(h(\alpha)p(\alpha|p,q,r,s)\right)$ and $\alpha_0$ is the point at which $u(\alpha)$ is maximized.
    \end{flushleft} 
    
    In the case of unnormalized density $q(\alpha|p,q,r,s)$, Laplace's method can be applied separately to $hq$ and $q$ to evaluate the numerator and denominator here :
    
    \begin{equation*}
    	\mathbb{E}[h(\alpha]=\frac{\int h(\alpha)q(\alpha|p,q,s,r)d\alpha}{\int q(\alpha|p,q,s,r)d\alpha}\;.
    \end{equation*}

\section{CONCLUSION}
\label{sec:conclusion}    

In this paper, we develop a mixture model to classify and cluster  partially observed data having outlier values. Hence a  scale mixture of Normal distributions, known for its robustness to outliers, is chosen. Moreover, thanks to the introduction of a latent variable, the proposed model can  model and infer on missing data. Model learning is processed through a Variational Bayes inference where a variational posterior distribution is proposed for missing values.

\bibliographystyle{plain}
\bibliography{refs}

%








\end{document}